\pdfoutput=1

\documentclass[11pt]{article}

\usepackage[final]{naacl2021}

\usepackage{times}
\usepackage{latexsym}
\usepackage{amsmath,graphicx}
\usepackage{bbm}

\usepackage[T1]{fontenc}

\usepackage[utf8]{inputenc}

\usepackage{microtype}

\title{Neural Model Robustness for Skill Routing in Large-Scale Conversational AI Systems: A Design Choice Exploration}
\author{
    Han Li, Sunghyun Park, Aswarth Dara, Jinseok Nam, Sungjin Lee, Young-Bum Kim, \\
    {\bf Spyros Matsoukas,} \and {\bf Ruhi Sarikaya} \\
    Amazon Alexa AI \\
    \small{\{lahl, sunghyu, aswardar, jinseo, sungjinl, youngbum, matsouka, rsarikay\}@amazon.com}
}

\begin{document}
\maketitle
\begin{abstract}
Current state-of-the-art large-scale conversational AI or intelligent digital assistant systems in industry comprises a set of components such as Automatic Speech Recognition (ASR) and Natural Language Understanding (NLU). For some of these systems that leverage a shared NLU ontology (e.g., a centralized intent/slot schema), there exists a separate skill routing component to correctly route a request to an appropriate skill, which is either a first-party or third-party application that actually executes on a user request. The skill routing component is needed as there are thousands of skills that can either subscribe to the same intent and/or subscribe to an intent under specific contextual conditions (e.g., device has a screen). Ensuring model robustness or resilience in the skill routing component is an important problem since skills may dynamically change their subscription in the ontology after the skill routing model has been deployed to production. We show how different modeling design choices impact the model robustness in the context of skill routing on a state-of-the-art commercial conversational AI system, specifically on the choices around data augmentation, model architecture, and optimization method. We show that applying data augmentation can be a very effective and practical way to drastically improve model robustness.
\end{abstract}

\section{Introduction}
\label{sec:intro}
\begin{figure}
    \centering
    \includegraphics[width=0.8\columnwidth]{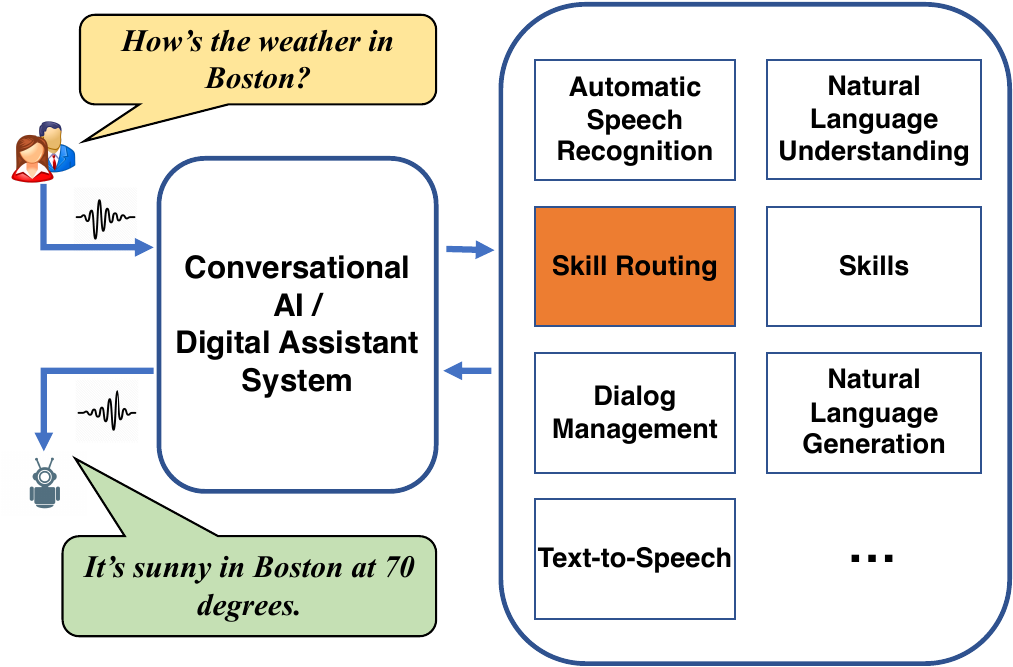}
    \caption{A high-level overview of different components in today's state-of-the-art large-scale conversational AI or digital assistant systems.}
    \vspace{-5mm}
    \label{fig:system_arch}
\end{figure}

Current state-of-the-art large-scale conversational AI or digital assistant systems in industry, such as Apple Siri, Amazon Alexa, Google Assistant, and Microsoft Cortana, involve a complex interplay of multiple components \cite{sarikaya2017technology} as shown in Figure \ref{fig:system_arch}. Here, we refer to skills to mean the applications that actually execute a user request, which comprise both first-party applications managed by in-house domains (such as Music or Weather) \cite{el2014extending} and third-party applications built and integrated by external developers (such as Uber or Spotify). Unless a user request goes through a special scenario such as multi-turn \cite{shi2015contextual}, contextual carry-over \cite{chen2016end}, or hypothesis re-ranking at a far downstream stage \cite{Robichaud2014HypothesesRF}, in some systems, routing the request to a skill is largely determined at the domain/skill classification step in the Natural Language Understanding (NLU) component. For other systems in which a shared ontology is used (e.g., a centralized intent/slot schema that can be leveraged by all domains/skills), there exists a separate skill routing component to correctly route a request to an appropriate skill after the semantic interpretation of the request in the NLU layer. The skill routing component is needed as there are hundreds or even thousands of skills and some of them either subscribe to the same intent in the schema (i.e., one-to-many mapping) and/or subscribe to certain intents only under a specific set of contextual conditions (e.g., device is headless or has a screen), which are external to NLU.

For the systems that leverage a centralized ontology, ensuring model robustness or resilience in the skill routing component is an important problem since skills, which are owned by different internal teams or external third-party developers, may dynamically change their subscription in the ontology after the model has been deployed to production. The model robustness in this context means the ability for the model to maintain its performance consistency even when the nature of the input to the model changes since training time - more specifically, the list as well as order of the skills coming into the model to be ranked. For instance, a skill can expand its functionality and newly subscribe to more intents (e.g., a \emph{Music} skill subscribing to more intents such as \emph{Play Meditation Sound}) or it can split into multiple skills with the original skill's scope of intents reduced (e.g., a \emph{Video} skill split into multiple skills such as \emph{YouTube Video} and \emph{TV Video} skills with a subset of \emph{Video}'s original list of intents). Addressing this problem is becoming even more important in the current environment where there is an explosive growth of third-party skills with many mainstream commercial conversational AI systems providing convenient toolkits for the third-party developers to quickly build and integrate new skills.

In this paper, we show with an extensive set of experiments how different modeling choices impact the model robustness in the context of skill routing (i.e., a model ranking a list of skills as routing candidates) on a state-of-the-art conversational AI system in production, specifically on the choices around data augmentation, model architecture, and optimization method. Data augmentation is a widely used technique that has shown to effectively improve model robustness across many tasks, including image classification \cite{krizhevsky2012imagenet}, automatic speech recognition \cite{park2019specaugment, cui2015data}, text classification \cite{wei2019eda}, reading comprehension \cite{wei2018fast}, language modeling \cite{dataaug:xie}, and task-oriented language understanding (through adversarial training) \cite{Einolghozati2019ImprovingRO}. On the model architecture, specifically for encoding context for list-wise ranking, we show the results of exploring two popular model architectures widely adopted in both industry and academia, Bi-LSTM \cite{schuster1997bidirectional} and Transformer \cite{vaswani2017attention}, as well as a simple aggregation-based architecture. For the optimization method, we show the results of exploring two different loss functions based on cross-entropy.

To our knowledge, this work is the first in the literature to introduce and extensively analyze the model robustness problem in the context of skill routing for a large-scale commercial conversational AI or digital assistant system. We introduce an approach of applying data augmentation for training the skill routing model and show that it can be a very effective and practical way to drastically improve model robustness in production. We also show that using the Bi-LSTM architecture to capture context for list-wise ranking can hurt model robustness, which still can be largely mitigated through data augmentation.

\section{Problem Definition}
\label{sec:prob_def}

For the skill routing model, given a dataset $\mathcal{D} = \{d_1, d_2, \dots, d_n\}$, each instance $d_i = (H_i, g_i)$ is a pair with $H_i = \{h_1, h_2, \dots, h_{|d_i|}\}$ being a list of hypotheses and $g_i \in H_i$ the ground-truth. Each $h_j \in H_i$ is a tuple $(u_j, n_j, s_j, c_j)$, where $u_j$ is the ASR-transcribed user request text, $n_j$ is an NLU interpretation (semantic understanding of the utterance text in terms of intent and slots), $s_j$ is a subscribed/proposed skill, and $c_j$ is a list of contextual signals such as whether the device has a screen. For simplicity, let's assume $f_m(\cdot)$ is a ranking model that outputs the top predicted hypothesis (thus, associated skill) from a list of hypotheses. We split $\mathcal{D}$ into the training set $\mathcal{D}_{train}$ and the test set $\mathcal{D}_{test}$, and train $f_m$ on $\mathcal{D}_{train}$ and evaluate offline on $\mathcal{D}_{test}$, with the test set accuracy defined as:
\vspace{-1mm}
\begin{equation*}
    Acc(f_m, \mathcal{D}_{test}) = \sum_{(H_i, g_i) \in \mathcal{D}_{test}} \frac{\mathbbm{1}[f_m(H_i) = g_i]}{|\mathcal{D}_{test}|}
    \vspace{-1mm}
\end{equation*}

In production, given a user request, a separate module called \emph{Hypothesis Proposer} is responsible for hypothesis list generation using the interpretation and the set of skills subscribed to the NLU interpretation (mostly on intents, but sometimes on a combination of intent and slots). At runtime, \emph{Hypothesis Proposer} also incorporates the dynamic updates of skill subscription to the shared ontology, hence can cause discrepancies on hypothesis generation during model training time and model service time: for an online user request $u_{on}$ that is semantically identical to an already observed $u_{off}$ in $\mathcal{D}_{train}$, the corresponding hypotheses $H_{on}$ and $H_{off}$ can be different. The possible transformation from $H_{off}$ to $H_{on}$ can be decomposed into two types of operations:

\vspace{-2mm}
\begin{itemize}
    \item Hypothesis removal ($H_{off}\setminus H_{on}$). The hypotheses no longer possible with the request are removed.
    \vspace{-2mm}
    \item Hypothesis insertion ($H_{on} \setminus H_{off}$). The hypotheses newly made possible are inserted into the hypothesis list.
\end{itemize}
\vspace{-2mm}

We explore a set of model variants $\mathcal{M}= \{f_{m_1}, f_{m_2}, \dots, f_{m_k}\}$ that implement various modeling design choices that can impact model robustness in the context of skill routing, and then evaluate the performance of each model $f_{m_i} \in \mathcal{M}$ using a state-of-the-art large-scale conversational AI system in production.

\section{Modeling Design Choice Exploration}
\label{sec:design}

\begin{figure*}[t]
\includegraphics[width=\textwidth]{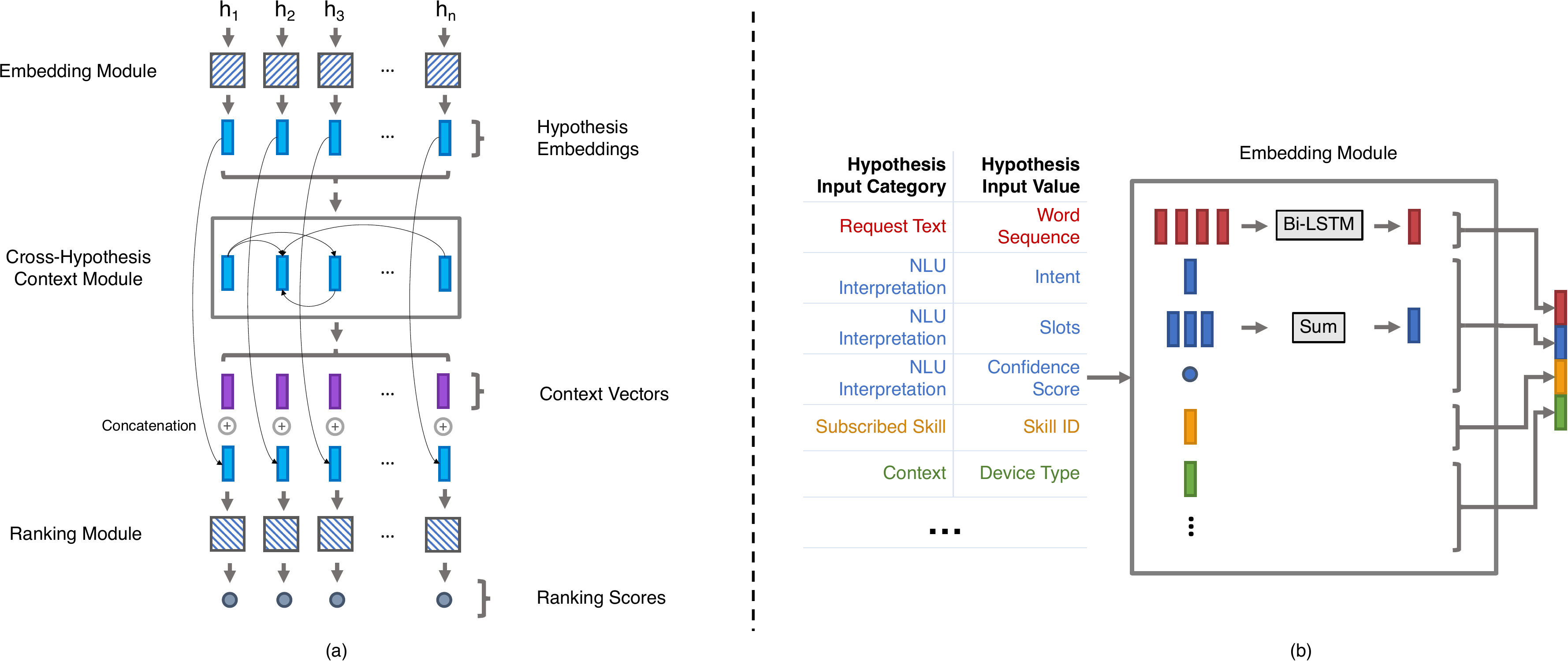}
\centering
\caption{(a) shows the base model architecture for skill routing and (b) shows a hypothetical embedding module that builds a hypothesis to be ranked by the model.}
\label{fig:model_arch}
\end{figure*}

This section describes three possible modeling design choices that can impact model robustness: model architecture, data augmentation, and optimization method.

\subsection{Base Model Architecture}
\label{subsec:model_arch}
Figure \ref{fig:model_arch}.(a) shows our base model architecture for skill routing, which consists of three neural network modules: \emph{embedding}, \emph{cross-hypothesis context}, and {\emph{ranking}}. Given a list of hypotheses $H = \{h_1, h_2, 
\dots, h_n\}$ of a user request where each $h_i = (u_i, n_i, s_i, c_i)$ defined in Section \ref{sec:prob_def}, the embedding module first converts each hypothesis $h_i$ into a fixed-size embedding vector $e_i$. Figure \ref{fig:model_arch}.(b) shows the composition of each hypothesis and the implementation detail of the embedding module. Once we have the hypothesis embeddings $e_1, e_2, \dots, e_n$, the cross-hypothesis context module takes as input all those embeddings and encodes contextual information across the hypotheses and then generates a context vector $c_i$ for each hypothesis $h_i$. Next, we concatenate the hypothesis embedding $e_i$ and the corresponding context vector $c_i$ of each hypothesis as the final representation $r_i = e_i \oplus c_i$. Finally, the ranking module, which is a two-layer feed-forward neural network, takes as input each $r_i$ to produce a ranking score $o_i$ that represents how likely $h_i$ can successfully handle the request. The hypothesis $h^*=\arg\max_{h_i\in H} o_i$ is top-ranked, and the skill associated with the hypothesis serves the request.

\subsection{Model Architecture Choices}
\label{subsec:model_choice}
Our exploration on model architecture choices focuses on the context module for two reasons. First, the implementation for the embedding and ranking modules are standard that are commonly used in practice. Second, the context module encodes the contextual information across the hypotheses, and therefore is more sensitive to changes in the possible hypothesis list. We explore three different options for the context module. The first option is a sequence-based architecture, where we use Bi-LSTM \cite{schuster1997bidirectional} for implementation. The second one is an attention-based architecture, and we use Transformer \cite{vaswani2017attention} for implementation. Both model architectures are widely adopted in many applications today \cite{young2018recent,devlin2018bert}. We also provide a simple aggregation-based method where we generate the context embedding by averaging all hypothesis embeddings as a baseline.

\paragraph{Sequence-based:} A Bi-LSTM \cite{schuster1997bidirectional} encodes the cross-hypothesis information in forward and backward directions to generate the context vectors. Given the hypothesis embeddings $e_{1:n}$ from the embedding module, a \emph{forward} LSTM processes the embeddings in the regular order to generate hidden state vectors $\overrightarrow{h}_{1:n}$, and a \emph{backward} LSTM processes the embeddings in the reverse order to generate $\overleftarrow{h}_{n:1}$.
The final context vector $c_i = \overrightarrow{h}_i \oplus \overleftarrow{h}_i$  is the concatenation of $\overrightarrow{h}_i$ and $\overleftarrow{h}_i$ for the $i$-th hypothesis. The order of the hypotheses influences the generation of the context vectors with this architecture.

\paragraph{Attention-based:} Given the hypothesis embeddings $e_{1:n}$, we first attend $e_i$ to each of all embeddings with an attention function $g$. The resulting attention score $s_{ij} = g(e_i, e_j)$ indicates the affinity of $e_i$ and $e_j$. Then, the context vector $c_i$ for the $i$-th hypothesis is generated by the weighted sum: $c_i =\sum_{j} (\exp{(s_{ij})}/\sum_{k}\exp{(s_{ik})})\cdot e_j$, representing a summary of context in $e_{1:n}$ that correlates with $e_i$. We use Transformer \cite{vaswani2017attention} as the implementation for the attention-based model architecture for capturing cross-hypothesis context. Compared with Bi-LSTMs, the attention-based architecture does not rely on the ordering information of the hypotheses, which can potentially be more robust to the changes in the hypothesis list.

\paragraph{Aggregation-based:} This is a baseline architecture in which we aggregate the hypothesis embeddings to encode the cross-hypothesis context information. Specifically, we directly take the average of the hypothesis embeddings as the context embedding. Compared to the first two options, this is a non-learning implementation built on heuristics, and therefore has the least modeling capacity. It also does not rely on any ordering information of the hypotheses and can potentially have less impact to the changes in the hypothesis list.

\subsection{Data Augmentation}
\label{subsec:data_aug}
Data augmentation is particularly suited for improving model robustness in our problem context since it simulates dynamic skill subscription during model training time. For each instance $(H_i, g_i) \in \mathcal{D}_{train}$, we augment $H_i$ by randomly injecting noise as follows. First, we get all NLU interpretations $N_i = \{n_j \mid n_j \in h_j \land h_j \in H_i\}$. Then, all hypotheses of $H_i$ with the same interpretation are grouped together, with each group denoted as ${H}_{n_j} = \{h \mid n \in h \land n_j \in N_i \land n=n_j\}$. Next, for each $n_j \in N_i$, we randomly generate $m$ hypothesis noise $\hat{H}_{n_j}$ without duplication. Each hypothesis noise $\hat{h}_{n_j,k} \in \hat{H}_{n_j}$ is generated with three steps. First, a random hypothesis $h_q \in H_{n_j}$ is selected. Suppose $\mathcal{S}$ is the space of all known skills, then we randomly select $\hat{s} \in \mathcal{S}$ with the condition that $\hat{s}$ is not the proposed skill of any hypothesis in ${H}_{n_j}$. Finally, we get $\hat{h}_{n_j,k}$ by replacing the proposed skill of $h_q$ with $\hat{s}$. Once we generate all hypothesis noise, we merge them all with $H_i$ to get the final augmented set $\hat{H}_i = \bigcup_{n_j\in N_i} \hat{H}_{n_j} \cup H_i$. By injecting noise into the training data, we expect the learned model to generalize better on unseen hypothesis combinations and more robust to hypothesis changes that occur after the model training time (i.e., while deployed to production without model retrain). 

In this work, we augment the training data by specifically injecting random noise at the hypothesis generation step. It reflects our expectation that new skills can be proposed with a uniform distribution given the limited information on how skill subscription to the shared ontology would change in the future, which has been shown to be effective in our experiment results (see Section \ref{sec:eval}). We note that there are other approaches possible. For example, we can perform more informed noise injection if we have prior knowledge that hypothesis generation would change in a pre-defined distribution or pattern based on historical data. We leave this exploration path as future work.

\subsection{Optimization Choices}
We also explore along the optimization axis. Specifically, we examine two commonly used loss functions: \emph{binary cross-entropy} (BCE) and \emph{multi-class cross-entropy} (MCE).

\paragraph{BCE:} Given a list of hypotheses $H_i = \{h_1, h_2, \dots, h_n\}$ of a user request, if we treat the ranking task for skill routing as a binary classification task (i.e., whether each $h_j \in H_i$ can be the invoked hypothesis or not), then the BCE loss function can be applied. Assume $O_i = \{o_1, o_2, \dots, o_n\}$ are the model prediction scores for $h_1, h_2, \dots, h_n$ respectively, and $h_k \in H_i$ is the ground-truth. Then the BCE loss is calculated as follows:
\vspace{-2mm}
\begin{equation*}
    \mathcal{L}_{}(O_i, h_k) = -\frac{1}{n}[\log(\sigma(o_k)) + \sum_{j\ne k} \log(1-\sigma(o_j))]
    \vspace{-2mm}
\end{equation*}
where $\sigma(\cdot)$ indicates the sigmoid function.

\paragraph{MCE:} We can also consider the ranking task as a multi-class classification problem. Under this setup, the MCE loss can be used which is calculated as follows:
\vspace{-2mm}
\begin{equation*}
    \mathcal{L}_{MCE}(O_i, h_k) = -o_k + \log(\sum_{j} \exp(o_j))
\vspace{-2mm}
\end{equation*}
Compared with BCE, MCE optimizes the objective with a more holistic view of the hypothesis list, and we expect the model trained with MCE to be more sensitive to changes in the hypothesis list.

\section{Results and Discussion}
\label{sec:eval}
\begin{table}[t]
\includegraphics[width=0.9\columnwidth]{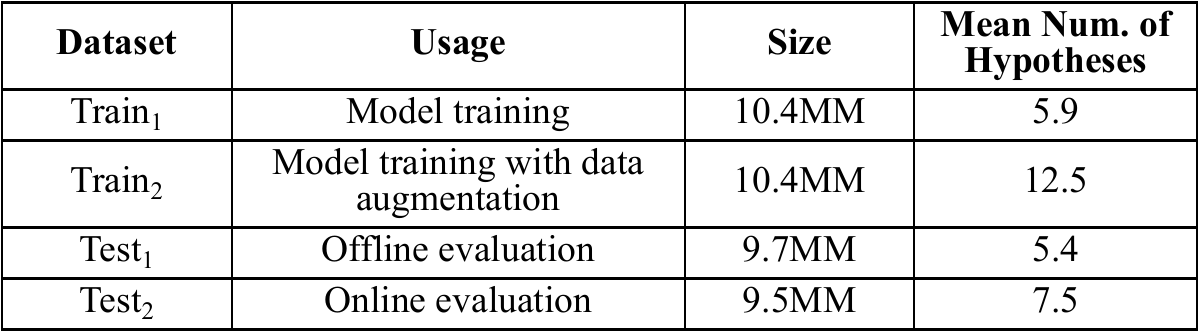}
\centering
\caption{Dataset statistics for the experiments.}
\label{table:datasets}
\end{table}

\paragraph{Dataset:} Table \ref{table:datasets} shows the statistics on the four datasets used for our experiments. Train$_1$ and Test$_1$ are each randomly sampled production traffic from non-overlapping time period and from one of today's state-of-the-art large-scale conversational AI systems. Train$_2$ is generated from Train$_1$ with data augmentation as described in Section \ref{subsec:data_aug}, where we select $m=3$ (sensitivity analysis showed little online performance gain with $m>3$). Test$_2$ is a random sample from another non-overlapping time period of production traffic emulating skill subscription changes in production\footnote{We will disclose further information on the dataset after confidentiality issues are addressed.}.

\vspace{-2mm}
\paragraph{Experiment Setup:} We train 12 model variants in total by combining different modeling design choices (see Section \ref{sec:design}), and evaluate each on Test$_1$ and Test$_2$ for offline and online evaluations to measure model robustness. All models are implemented in PyTorch \cite{paszke2019pytorch} and trained and evaluated with Intel Xeon E5-2686 CPUs, 244 GB memory, and 4 NVIDIA Tesla V100 GPUs. We use Adam \cite{kingma2014adam} as the optimization algorithm for all model training for 15 epochs with 1024 batch size.

\subsection{Overall Evaluation}
\label{subsec:overall_eval}

Table \ref{table:overall_eval} shows the overall evaluation on skill routing  accuracy of the 12 model variants on Test$_1$ (i.e., offline) and Test$_2$ (i.e., online). Here, we use the offline accuracy of the model with Bi-LSTM with BCE loss as the baseline, denoted as $s_b$, and we show the performance differences against the baseline (in percentage) for all other model variants\footnote{We currently cannot disclose the absolute performance metrics at this moment of paper submission, but for better interpretation of our experiment results, we note that the offline accuracy for the baseline model $s_b$ with Bi-LSTM and BCE loss is >90\%. We will disclose full detail after the confidentiality issues are addressed.}. From the table, we observe the following.

First, in terms of offline evaluation on Test$_1$, the attention-based model achieves the best overall performance with 1.49\% (with data augmentation) and 0.93\% (no data augmentation) accuracy improvement over the baseline $s_b$ when BCE is used, both of which are higher than the Bi-LSTM model's (i.e., 1.16\% and 0\%) and the aggregation-based model's (i.e., 1.10\% and -0.04\%). When MCE is used, Bi-LSTM achieves the best overall performance with 1.19\% and 0.10\% accuracy improvement. The aggregation model consistently produces the worst accuracy results under both scenarios, indicating that a higher model capacity can help achieve higher offline model performance. 

\begin{table}[t]
\includegraphics[width=\columnwidth]{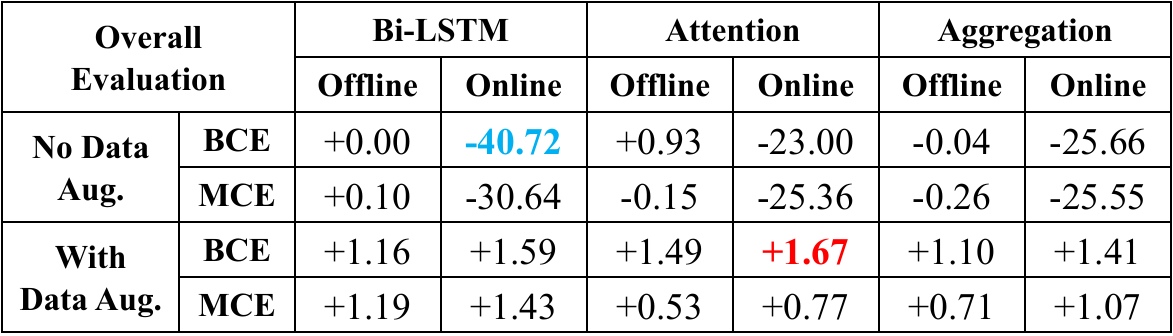}
\centering
\caption{The overall evaluation on the skill routing accuracy (in percentage) across the 12 model variants. The model variant trained with Bi-LSTM and BCE without data augmentation is the baseline, and all other model performance shown is the absolute difference against the baseline (e.g., if the baseline is at 100\% accuracy and one model variant shows -5.00\%, its accuracy would be 95\%.}
\label{table:overall_eval}
\end{table}

\begin{table}[t]
\includegraphics[width=\columnwidth]{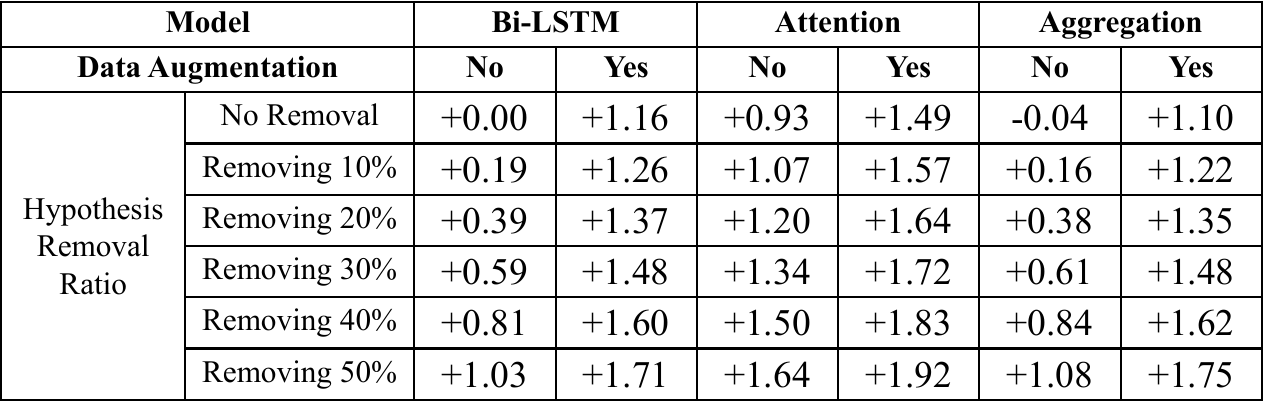}
\centering
\caption{Skill routing accuracy as we vary the level of hypothesis removal.}
\vspace{-5mm}
\label{table:hyp_removal}
\end{table}

Second, when we evaluate the model variants on Test$_2$, the accuracy scores diverge drastically. For instance, for the Bi-LSTM model using BCE without data augmentation, the online accuracy drops 40.72\% compared to its offline score $s_b$, while the attention-based model using BCE with data augmentation gain 1.67\% accuracy improvement over $s_b$. This indicates that even with similar offline evaluation performance, different model architecture designs can have a big impact on model robustness in an online setting where skill subscription can change dynamically after model deployment.

Third, by freezing the data augmentation choice and loss function to only compare different model architecture designs on Test$_2$, we see that Bi-LSTM produces the worst performance. Specifically, when disabling data augmentation, Bi-LSTMs has 40.72\% and 30.64\% accuracy drop on BCE and MCE respectively, which are much lower than the attention-based architecture (23.00\% and 25.36\% drop) and aggregation-based architecture (25.66\% and 25.55\% drop). This suggests that the Bi-LSTM model, which is sensitive to the ordering of hypotheses, is more susceptible to potential hypothesis changes at runtime, and most likely not a good design choice. On the other hand, the attention model achieves the best online evaluation performance, indicating the best choice among the three.

Fourth, comparing the two loss functions by freezing the other two types of design choices, we see that BCE outperforms MCE in four cases out of six for both offline and online evaluations, indicating that BCE is likely a better optimization method in general in the context of our skill routing problem.

Last, for all models with data augmentation, we observe no online accuracy degradation. We observe slightly higher accuracy results than the offline performance in all cases. One possible reason can be that the model training with noise injection helps the model to be more generalizable, especially to the changes in the list of hypotheses. Also, some hypotheses that originally exist in the hypothesis list can be removed at model runtime due to skills dynamically unsubscribing to the shared ontology, reducing the number of competitive hypotheses making the ranking task easier.

\begin{table}[t]
\includegraphics[width=\columnwidth]{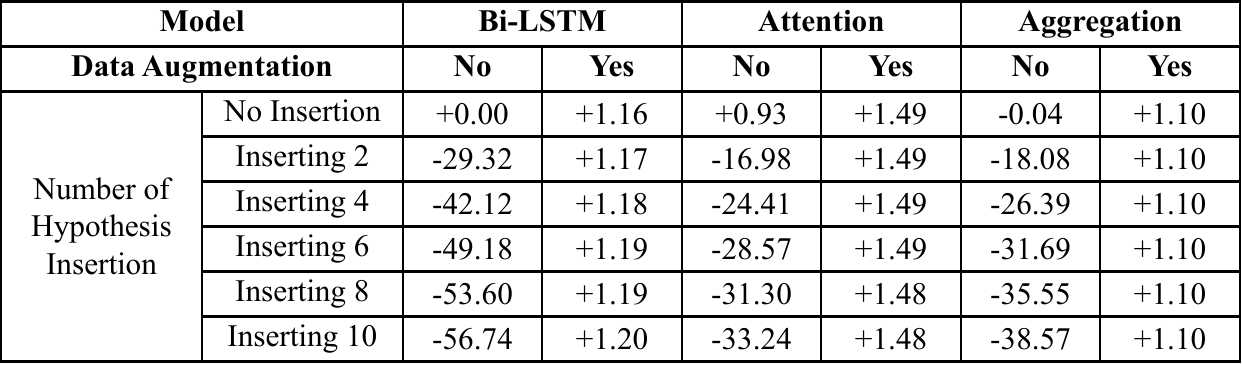}
\centering
\caption{Skill routing accuracy as we vary the level of hypothesis insertion.}
\vspace{-6mm}

\label{table:hyp_insert}
\end{table}

\vspace{-2mm}
\subsection{Micro-Benchmarks}
We show micro-benchmarks to evaluate the impact of hypothesis removal and insertion separately. We fix BCE as the loss function and present results on the six model variants, noting that the results using MCE also share similar patterns. Similarly to the overall evaluation in Section \ref{subsec:overall_eval}, we use the offline accuracy $s_b$ with Bi-LSTM and BCE as the baseline (i.e., top-left score in Table \ref{table:hyp_removal} and \ref{table:hyp_insert}) and show the performance changes against the baseline (in percentage) for the other model variants.

\emph{Hypothesis Removal}: We modified Test$_1$ by randomly removing a subset of non-golden hypotheses of each test instance, to examine how different modeling design contribute to the robustness with respect to hypothesis removal online. Specifically, we varied the removal ratio from 0 to 50\%, as Table \ref{table:hyp_removal} shows. As we increase the removal ratio, for all model variants, we observe a pattern of increasing accuracy. We believe the reason for this tendency is straight-forward: as more hypotheses are removed, there will be fewer candidates for the model to rank, which makes the routing task easier.

\emph{Hypothesis Insertion}: We modified Test$_1$ by randomly inserting new hypotheses to the hypothesis list of each instance. Specifically, we varied the number of insertions from 0 to 10, as shown in Table \ref{table:hyp_insert}. First, when data augmentation is not applied, as we increase the number of insertions, the accuracy scores for all model variants decrease drastically. This means compared to hypothesis removal, hypothesis insertion has a much bigger and adverse impact on model performance. Second, the accuracy on Bi-LSTM drops by 56.74\% when inserting 10 hypotheses to each test instance in Test$_1$, which is much lower than the other two model architecture choices, suggesting the worst robustness capability. Finally, after applying data augmentation, all models achieve similar prediction accuracy to the offline evaluation results. This indicates regardless of other design choices, data augmentation can effectively address the model robustness problem induced by hypothesis insertion.

\section{Conclusion}
\label{sec:conclusion}
We investigated the model robustness problem in the context of skill routing in one of today's state-of-the-art large-scale conversational AI systems, specifically by exploring a set of modeling design choices from three perspectives: model architecture choice, data augmentation, and optimization method. With extensive experiments, we showed that applying data augmentation can be an effective and practical way to drastically improve model robustness. Also, using the Bi-LSTM architecture to capture cross-hypothesis context in a list-wise ranking fashion can hurt model robustness, which still can be largely mitigated with data augmentation.

\newpage
\bibliographystyle{acl_natbib}
\bibliography{mybib}

\end{document}